\documentclass[11pt]{article}
\usepackage[letterpaper,margin=0.95in]{geometry}
\usepackage{amsmath,amssymb,booktabs,flafter,graphicx,microtype}
\usepackage{capt-of}
\usepackage{xcolor}
\definecolor{externalblue}{HTML}{174E80}
\usepackage[colorlinks=true,urlcolor=externalblue,linkcolor=black,citecolor=black]{hyperref}
\newcounter{algorithm}
\newenvironment{methodsteps}[2]{\par\medskip\noindent\begin{minipage}{\linewidth}
  \refstepcounter{algorithm}\label{#1}\hrule\vspace{0.45em}\noindent\textbf{Algorithm \thealgorithm. #2}\par\smallskip
}{\vspace{0.25em}\hrule\end{minipage}\par\medskip
}
\graphicspath{{figures/}}
\title{Millie 2: Streaming a 177-Billion-Parameter Language Model on Phones at Million-Token Context}
\author{Ben Krause\\LLMs For All\\{\small ben.krause@llmsforall.com}}
\date{}

\begin{document}

\maketitle

\begin{abstract}
Millie 2 runs a 177-billion-parameter language model with a million-token context on a phone with 12\,GB of RAM. At that length, its compressed weights and attention data occupy 69.5\,GB. On an iPhone 18 Pro, it generates 14.1 tokens/s at short context and 9.9 tokens/s when answering a question about a million-token document. Millie 2 streams model weights by predicting which experts a layer will use and favoring experts whose weights are already in RAM. We propose Residency-Biased Attention (RBA) for streaming attention data. Millie 2 uses an earlier layer's state to score a compact index of the full history, giving a bonus to blocks whose keys and values are already in RAM. The model fetches missing blocks while earlier layers compute. The destination layer then computes attention with the per-token keys and values of its selected blocks. To recover model quality, we keep the compressed weights fixed and train their shared scales under the same expert-selection constraints used on the phone. The resulting model scores close to its parent across reasoning, coding, instruction-following, and agentic benchmarks. We also evaluate browser and Android agents with model inference on the phones. These methods make it practical to run a large language model at long context on a phone even when its compressed weights and attention history together far exceed the phone's RAM. We release the model weights and inference code.
\end{abstract}

\begin{quotation}
\footnotesize
\noindent\textbf{Code and model weights:}\enspace
\url{https://github.com/llmsforall/millie-runtime#millie-2}
\end{quotation}

\section{Introduction}

Much of the progress in language modeling has come from training larger Transformers on more data~\cite{transformer,chinchilla}. GPT-3 showed that a single model with 175 billion parameters could learn a task from a few examples in its prompt~\cite{gpt3}. Instruction tuning and reinforcement learning from human preferences have since made language models easier to direct with ordinary instructions~\cite{flan,instructgpt}. What we ask of a model has also changed. Answering a question about a document can require finding a passage hundreds of pages earlier. Fixing a bug can require reading several files, running tests, and using the results to revise a patch. Recent models have made substantial gains in solving mathematical problems, writing software, and using tools to carry out multistep tasks~\cite{glm5,qwencoder,kimi3,deepseek41}. The ability to work with a long input has become valuable in its own right, and models such as Gemini 1.5 have demonstrated it on inputs exceeding a million tokens~\cite{gemini15}.

For someone using a large language model on a phone, the model itself usually runs in the cloud. A provider can spread its weights across several accelerators and use the same weights to serve many people at once~\cite{vllm,deepseekv3}. For a personal assistant, the information needed to answer a question often lives on the phone. To explain why a project changed direction, for example, it might need the original proposal, the conversation in which the decision was made, and the revised plan. A long context lets the assistant consider those records together. Local inference lets it do so privately, even offline. To provide both, a phone has to accommodate the model and its growing attention history in the memory available to one application. This helps explain why Apple Intelligence, for example, pairs a three-billion-parameter model on the device with a larger model on its servers~\cite{applefm}.

How much of a large model actually needs to be in memory while it generates an answer? In earlier work, we ran Millie 1.1, a 35-billion-parameter model, on an iPhone by keeping some of its weights on storage~\cite{millie11}. Here we take on both sources of memory demand. Millie 2 adapts Qwen3.8-Flash-Next~\cite{qwen}, a model with 177 billion text parameters, to run on phones at up to a million tokens of context. Even after compression, our main checkpoint occupies 56.3\,GB. At a million tokens, the keys and values that attention uses to refer back to earlier text occupy another 13.1\,GB. The iPhone we use has 12\,GB of RAM in total.

What makes this possible is that the data needed to generate any one token is much smaller than either total suggests. Millie 2 uses about 6.7 billion parameters per token. At each layer, it routes the token through a few feed-forward networks, called experts, drawn from a much larger set. Its sparse attention layers likewise select portions of the input history according to their relevance to the current token~\cite{nsa,dsa}. A phone has enough flash storage to keep all these candidates available, and prior work has shown how to fetch selected weights from storage during inference~\cite{flash,eliseev}. The difficulty is timing. By the time a layer has selected an expert or a passage, it is ready to use the data, and a storage read can leave it waiting. We change how the model selects both kinds of data so it reuses what is already in RAM more often and starts the remaining reads while earlier layers are still computing. We also train the compressed model under the expert-routing restrictions it will encounter on the phone. This leads to three contributions.

\begin{enumerate}
\interlinepenalty=10000
\item \textbf{Efficient expert streaming.} We use an earlier layer to predict which expert weights will be needed, favor weights already in RAM, and start reads while the network is still computing. The destination layer then routes among the available experts using its own state.
\item \textbf{Residency-Biased Attention (RBA).} We keep a compact index of the full history in RAM and use it to choose which blocks of keys and values to retain or fetch. The selection favors blocks already in memory. The destination layer can score the available blocks again with its current query before applying attention.
\item \textbf{Log-parameterized scale training.} We recover quality by learning shared scales with the compressed weight codes held fixed, and train under the expert-routing restrictions used at inference. Our kernels use the packed weights during both the forward and backward passes.
\end{enumerate}

On an iPhone 18 Pro, Millie 2 generates 14.1 tokens/s at short context and 9.9 tokens/s in the 1M-context document test. It retains much of the parent model's performance on reasoning, coding, instruction-following, and agentic benchmarks. We also run model inference on phones for agents that complete tasks in a browser or Android environment. To test the hardware beyond the model's supported context length, we repeat a saved prefix and measure RBA decode at 7.6 tokens/s on the iPhone at 2.5M positions. These results show how a phone can handle workloads whose model weights and attention history would otherwise each exceed its RAM budget.

\section{Background}
\subsection{Parent model architecture}
\label{sec:architecture}
Qwen3.8-Flash-Next has 48 text-model layers, arranged as twelve repetitions of three Gated DeltaNet layers~\cite{gateddelta} and one Qwen Sparse Attention (QSA) layer. Each of the 48 layers follows its sequence-mixing operation with a mixture-of-experts module. The model uses a gated residual stream with four branches. Learned gates control how each operation reads from and writes to those branches \cite{qwen}. During autoregressive decoding, each Gated DeltaNet layer updates a recurrent state of fixed size as it reads each token. The twelve QSA layers retain the keys and values of earlier tokens, so their cache grows with sequence length. The combination keeps the recurrent part independent of context length and concentrates the growing cache in the sparse attention layers. For million-token inference, we use YaRN~\cite{yarn} with a context-extension factor of four for the rotary position embeddings, extending the configured window from 262,144 to 1,048,576 positions.

\paragraph{Parameter count and $n$-gram embeddings.}
The bigram and trigram embedding tables contain 51.2B parameters. Together with 125.7B parameters elsewhere in the text model, they bring the total to about 177B. The model indexes those tables using short sequences of adjacent tokens at the second layer. A token accesses only the rows indexed by its bigram and trigram identifiers.

\paragraph{Vision encoder.}
The parent model has a separate 0.45B-parameter vision encoder that converts images into embeddings used by the text model.

\begin{figure}[tbp]
\centering
\includegraphics[width=0.88\linewidth,height=0.47\textheight,keepaspectratio]{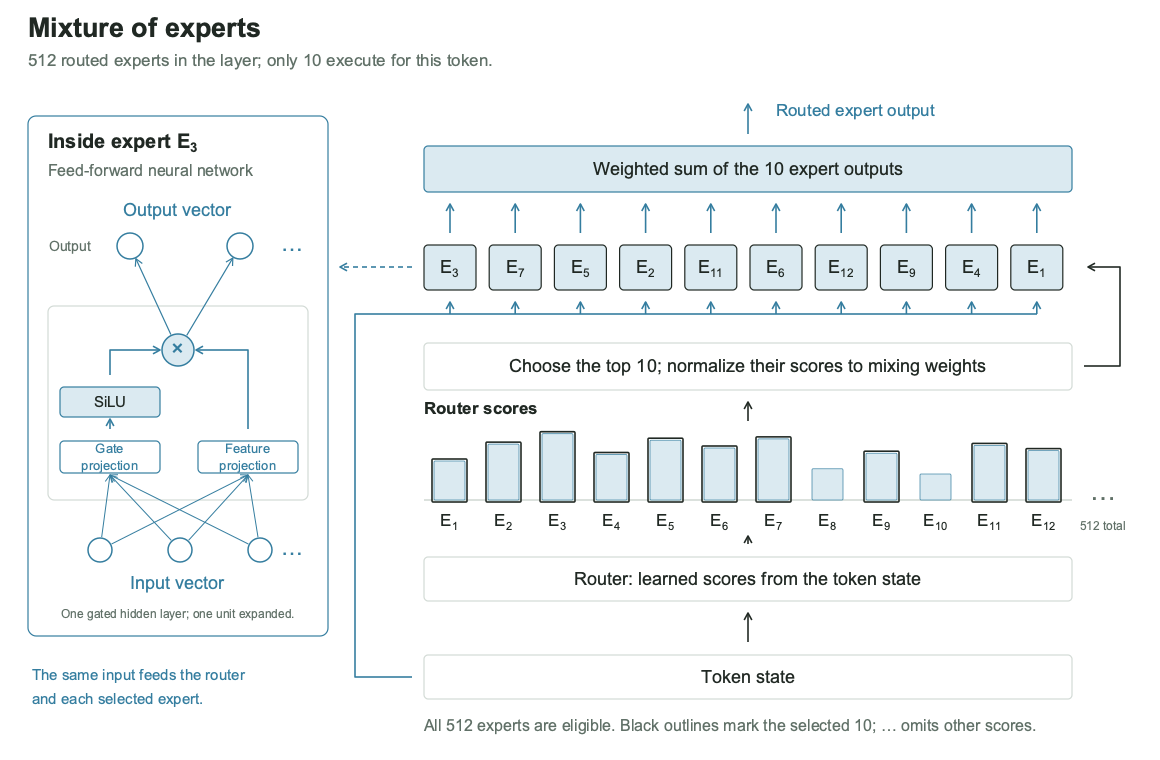}
\caption{The routed part of a mixture-of-experts module. Each selected expert receives the same token state, and the router's weights determine how their outputs are combined. The inset expands one gated feed-forward expert.}
\label{fig:moe-background}
\end{figure}

\paragraph{Mixture of experts.}
In a sparsely routed mixture of experts, a learned gate selects a few expert networks for each input and weights their contributions~\cite{jacobs,shazeer,mixtral}. Each mixture-of-experts module in our parent model has 512 routed experts and one shared expert. An expert is a gated feed-forward network that projects a token's hidden state to an intermediate representation and then back to the hidden dimension. A router scores the routed experts from the same token state, selects ten, and combines their outputs using the routing weights. The shared expert also contributes an output for every token. Figure~\ref{fig:moe-background} illustrates the routed path.

\begin{figure}[!t]
\centering
\includegraphics[width=0.94\linewidth,height=0.56\textheight,keepaspectratio]{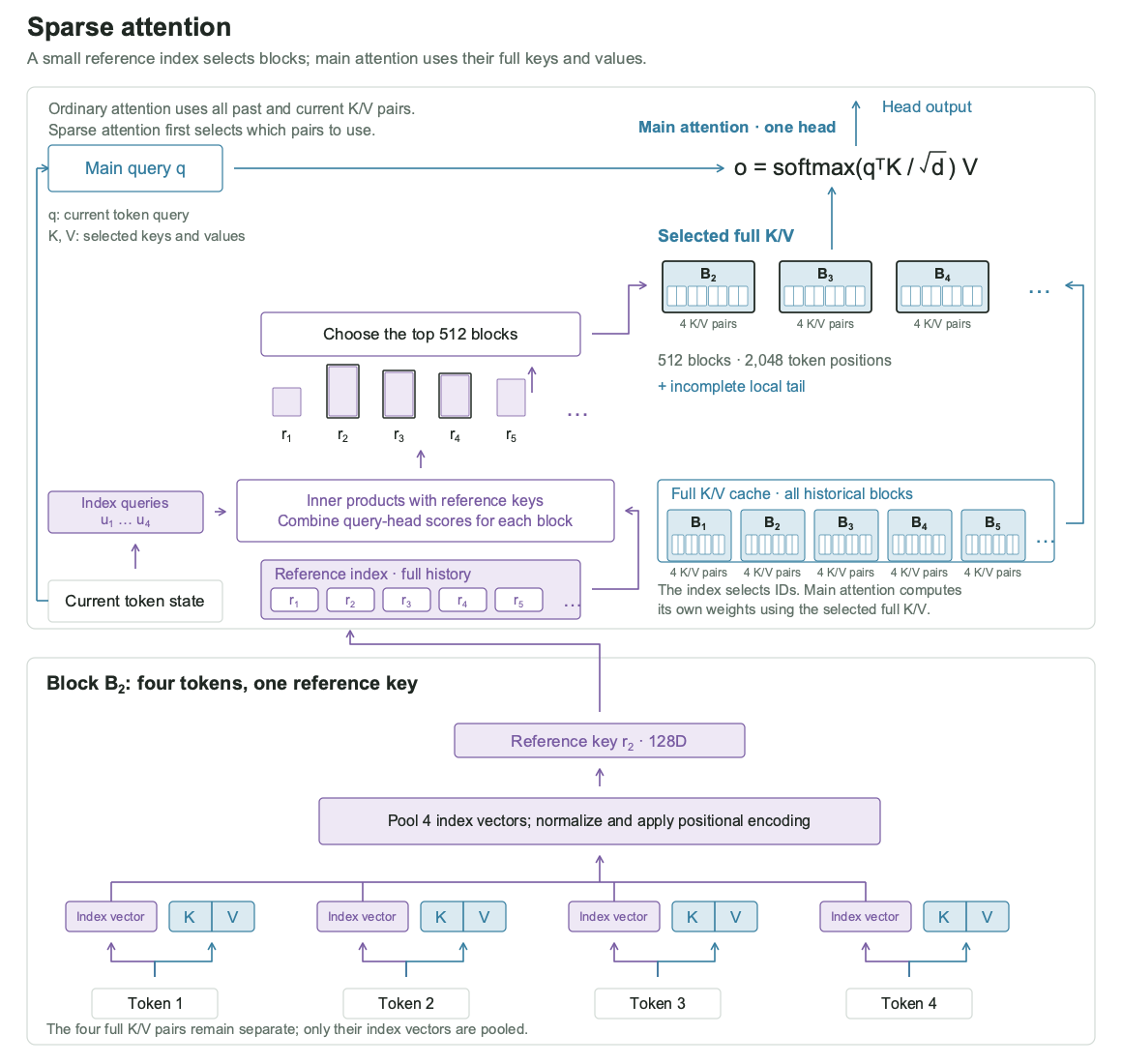}
\caption{Qwen Sparse Attention. The indexer scores compact reference keys to select blocks. The main attention computation uses the individual keys and values at the selected positions.}
\label{fig:sparse-attention-background}
\end{figure}

\paragraph{Qwen Sparse Attention.}
In dot-product attention, a query represents the current token and a key describes a position it could attend to. After normalization, the query--key scores weight the values in the output sum~\cite{luong,transformer}. Sparse attention computes this sum over selected positions, which can be specified by a fixed pattern or selected from the input~\cite{sparseformer,nsa,dsa}. QSA uses a separate block selector before computing those attention weights. Its indexer pools four consecutive index vectors into a 128-dimensional reference key and uses its own query vectors to score the keys and select up to 512 blocks from the history. Each token retains its own attention key and value. Main attention uses its own query on the selected tokens and the incomplete local block (Figure~\ref{fig:sparse-attention-background}). The native policy keeps attention data for all eligible blocks available for retrieval.

\section{Methods}

\subsection{Expert streaming}
\label{sec:expert}
We expand on the expert-streaming methods used in Millie 1.1~\cite{millie11}. Millie 2 uses ten routed experts per layer and makes two decisions at different times. An early prediction nominates ten experts whose weights should be available. Once the token reaches the destination layer, its router selects ten experts to run from the updated resident pool.

For layer $\ell$, let $a_{\ell,e}$ be the early score for expert $e$, $R_\ell$ the resident experts, and $P_\ell$ the highest-scoring nonresident candidates protected during admission. Let $R'_\ell$ be the pool after admission and $z_{\ell,e}$ the destination router's score from the current token state. The nominations $N_\ell$ and active experts $S_\ell$ are
\begin{align}
 N_\ell &= \operatorname{Top}_{10}\!\left(\bigl(a_{\ell,e}+b\,\mathbf{1}\{e\in R_\ell\cup P_\ell\}\bigr)_e\right), \label{eq:expert-admission}\\
 S_\ell &= \operatorname{Top}_{10}\!\left((z_{\ell,e})_{e\in R'_\ell}\right). \label{eq:expert-active}
\end{align}
Here $\operatorname{Top}_{10}$ returns expert IDs. The indicator adds the bonus $b$ to the early scores of residents and protected candidates. The destination router uses its own scores to select and weight the active experts. Since the pool can hold more than ten experts, a resident expert can run even when the early predictor did not nominate it.

\paragraph{Lookahead admission.}
For destination layer $\ell$, the predictor takes the residual state after the sequence-mixing operation in layer $\ell-1$, while the expert computation for that preceding layer remains to be done. The residual stream at this point has four branches. We read them with the expert-input mixer belonging to layer $\ell$: its learned gates weight the branches to form a single input vector $x_\ell^{\mathrm{early}}$ for the router. When execution reaches the expert module of layer $\ell$, we apply the same mixer parameters again, now to the updated residual branches after that layer's sequence-mixing operation. The gates depend on the residual state, so this second read gives the router a fresh input for final selection. We apply low-rank adaptation (LoRA)~\cite{lora} to the destination router. With router matrix $W_\ell$ and learned adapter matrices $A_\ell$ and $B_\ell$, the vector of predicted admission scores is
\begin{equation}
    a_\ell = W_\ell x_\ell^{\mathrm{early}} +
             B_\ell\!\left(A_\ell x_\ell^{\mathrm{early}}\right).
    \label{eq:expert-predictor}
\end{equation}
Appendix~\ref{app:lookahead} gives the residual-mixer equations for the early and destination reads. To form $P_\ell$ in Eq.~\ref{eq:expert-admission}, we take the $m$ highest-scoring nonresident experts, or all of them if fewer are available. Each expert in $P_\ell$ therefore competes with residents on its original predicted score. It is fetched if it ranks among the ten nominations. Other nonresident experts can also qualify if their predicted scores overcome the residency bonus. We usually set $m=1$ because the preceding layer's remaining computation can hide the read for one expert. The runtime starts reads for nominated experts absent from RAM, then evicts the least recently nominated residents outside the current nomination set as needed. It refreshes the recency of nominated experts. Experts outside the nomination set can remain in the pool when space allows.

\begin{figure}[tbp]
\centering
\includegraphics[width=0.89\linewidth,height=0.49\textheight,keepaspectratio]{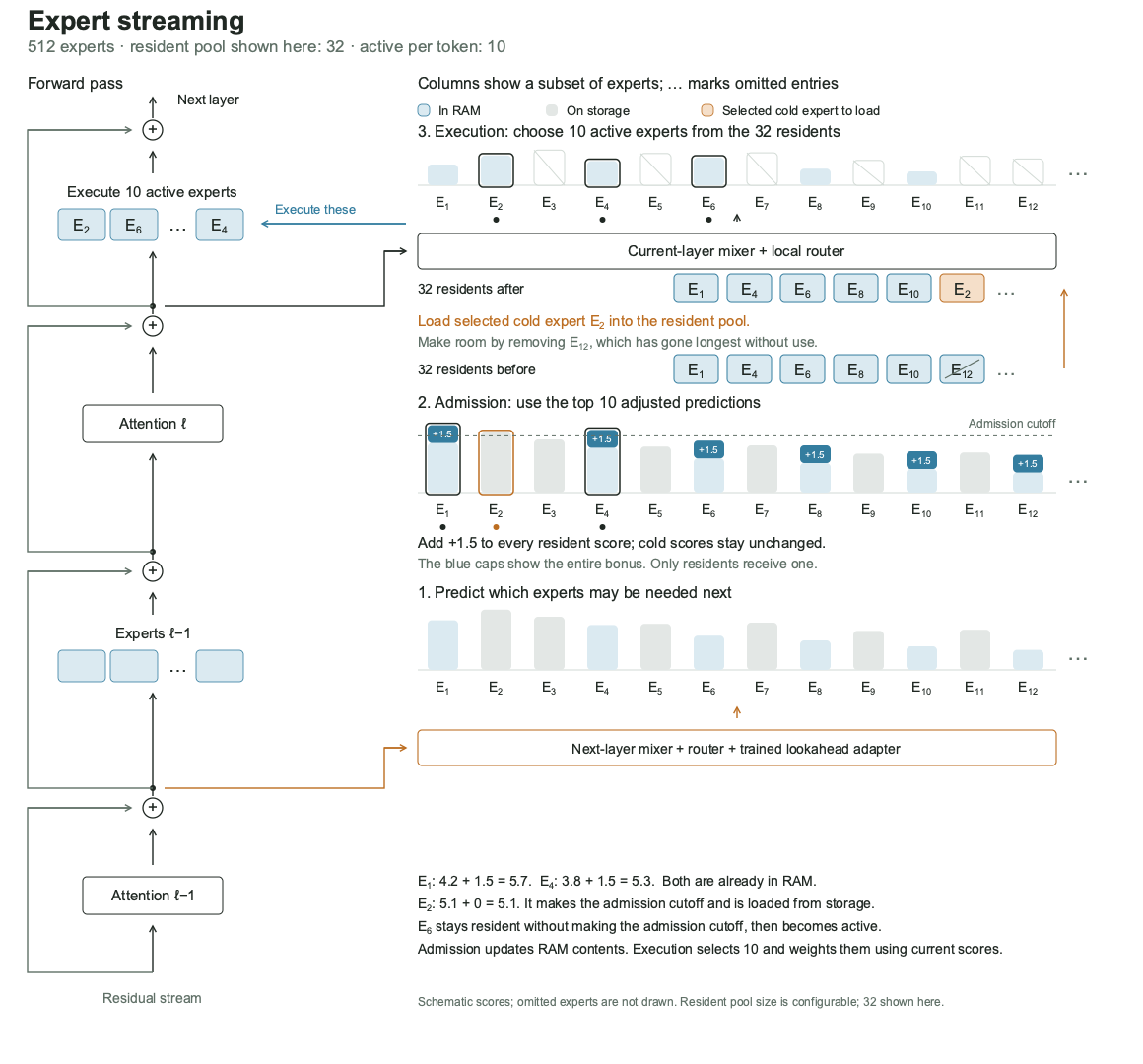}
\caption{Expert streaming. Early scores determine admissions to the resident pool. The destination layer selects active experts from that pool using its current hidden state.}
\label{fig:expert}
\end{figure}

\paragraph{Final selection.}
At the destination layer, the router scores the updated pool using the actual input to that layer's expert module. It chooses ten active experts from that pool and computes their mixture weights from the current routing scores. An expert retained from an earlier token can therefore become active even if the early predictor did not nominate it for this token. The shared expert still contributes at every token. Figure~\ref{fig:expert} and Algorithm~\ref{alg:expert} show both selection steps.

\paragraph{Training the lookahead router.}
We train lookahead routers for both the ternary and three-bit expert models, separately from quantization recovery. In the three-bit training run, we hold the model weights and router fixed and train only the rank-eight adapter matrices in Eq.~\ref{eq:expert-predictor}. At each completion position, the predictor uses the earlier state and learns to match the original model's routing distribution over all 512 experts at the destination layer. This training uses scores before the admission bonus is applied.

\begin{methodsteps}{alg:expert}{Expert streaming for one token at destination layer $\ell$}
\textbf{Inputs.} Resident pool $R_\ell$, early state from layer $\ell-1$, admission bias $b$, protected nonresident count $m$\par
\begin{enumerate}
\item Predict admission scores using Eq.~\ref{eq:expert-predictor} and select the top $m$ nonresident experts as $P_\ell$.
\item Use Eq.~\ref{eq:expert-admission} to nominate the ten highest-ranked experts.
\item Start asynchronous reads for nominated experts absent from RAM and continue computing layer $\ell-1$.
\item Install the fetched weights. If space is needed, evict the least recently nominated experts outside the current nominations.
\item Select ten experts from the updated resident pool using the destination layer's actual input.
\item Combine their outputs with the current routing weights and add the shared expert output.
\end{enumerate}
\end{methodsteps}

\subsection{Residency-Biased Attention}
\label{sec:attention}
The reference index stores one compact key for each completed historical block. We keep the index in RAM, put the full historical keys and values on storage, and retain a bounded per-layer pool of blocks whose keys and values are available to attention. Before the history exceeds the working pool, the runtime can use native selection over the locally available history. Once streaming begins, scoring the reference index can nominate blocks from any point in the history.

\paragraph{Admission and active selection.}
Let $I_\ell$ be the IDs of the completed historical blocks in layer $\ell$'s reference index and $R_\ell\subseteq I_\ell$ the IDs of those whose keys and values are resident. The destination layer's indexer projects an early query from an earlier layer state and produces an unadjusted relevance score $s_j$ for each indexed block $j$. For the streaming policies here, the residency-adjusted score is
\begin{equation}
    \widetilde{s}_j=s_j+\beta\sigma_s\,\mathbf{1}\{j\in R_\ell\},
    \qquad
    \sigma_s^2=\frac{1}{|I_\ell|}\sum_{j\in I_\ell}(s_j-\bar{s})^2 .
    \label{eq:attention-admission}
\end{equation}
Here $\bar{s}$ is the mean early score and $\beta$ sets the bias in units of one standard deviation. The top 512 adjusted scores nominate blocks for admission. The runtime can start reads for missing blocks while earlier layers are still computing. The residency adjustment acts on block admission. The main attention operation uses the destination layer's attention query and the per-token keys and values for its chosen blocks.

\begin{figure}[tbp]
\centering
\includegraphics[width=\linewidth,height=0.76\textheight,keepaspectratio]{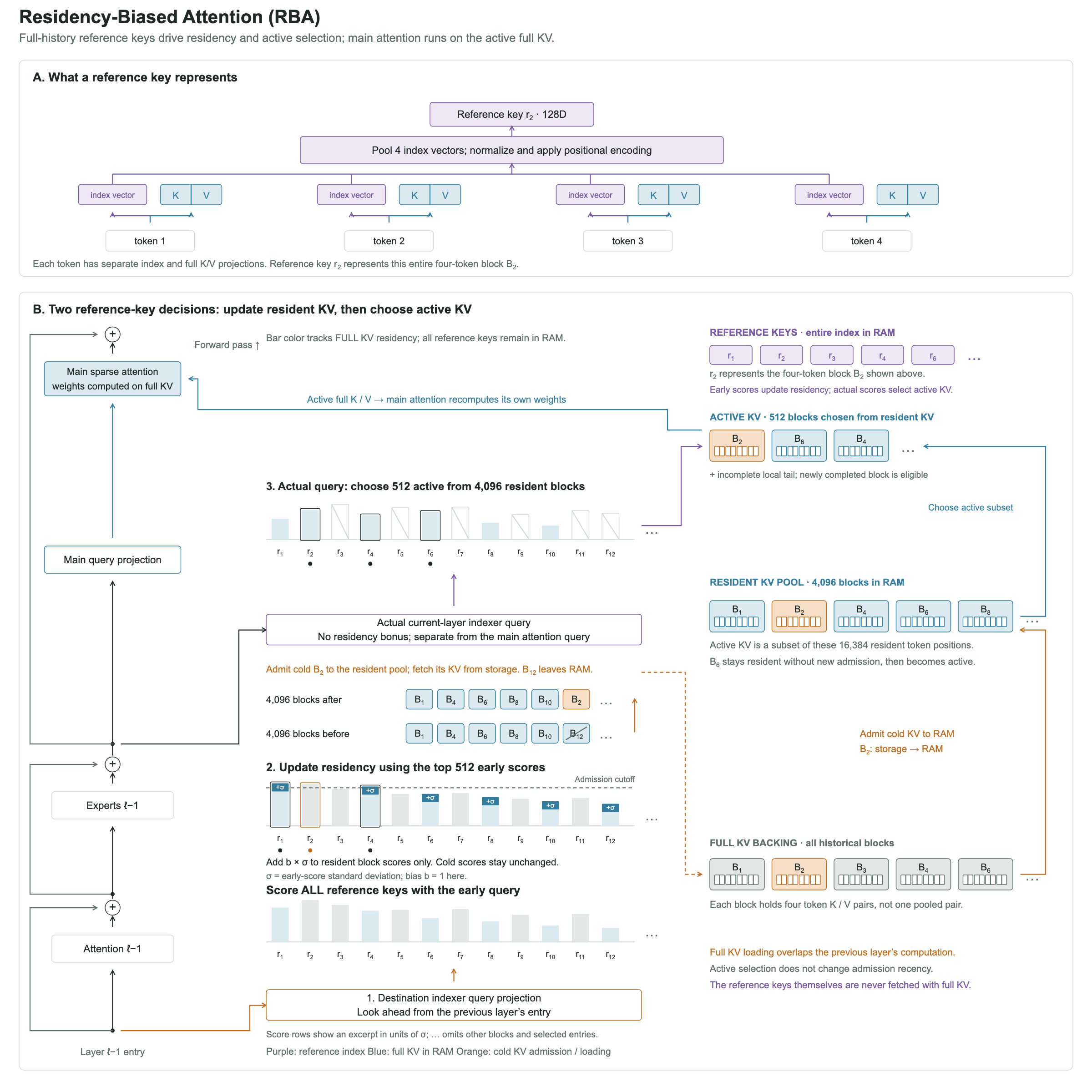}
\caption{Attention streaming with Residency-Biased Attention (RBA). The index covers the entire history. Early queries control residency of full KV blocks. In the reranking policy the current query chooses active blocks from the resident pool, and the selected full keys and values define the attention output.}
\label{fig:attention}
\end{figure}

\paragraph{Lookahead policies.}
The destination layer's attention-input mixer reads the four residual branches twice: first at the lookahead point and again at the destination. Its learned parameters are shared across the two reads, while its gates depend on the state being read. Appendix~\ref{app:lookahead} gives both reads, the indexer projections, and the resulting block scores. In the one-layer reranking policy, we form the early indexer query at the entry to layer $\ell-1$. Layer $\ell$ then derives a new indexer query from its current state, scores the eligible blocks in the updated pool without the residency adjustment, and selects up to 512 completed blocks. A newly completed block can also be eligible from local data. In the three-layer early-selection policy, we form the early query at the entry to $\ell-3$ and reuse the nominations for historical active selection. This gives the reads a longer interval to overlap with computation. In both policies the still-incomplete block can supply its locally available keys and values under the causal mask. Figure~\ref{fig:attention} illustrates the reranking path. Algorithm~\ref{alg:attention} shows the shared steps and the choice of active set.

\begin{methodsteps}{alg:attention}{Attention streaming with RBA for one token at destination layer $\ell$}
\textbf{Inputs.} Completed-block index $I_\ell$, resident pool $R_\ell$, bias multiplier $\beta$, lookahead distance $d$\par
\begin{enumerate}
\item Project an early indexer query from the state at entry to layer $\ell-d$.
\item Score $I_\ell$ and nominate $N\gets\operatorname{Top}_{512}(\widetilde s)$ using Eq.~\ref{eq:attention-admission}.
\item Start reads for nominated blocks absent from RAM and continue earlier-layer computation.
\item Install the fetched blocks in the bounded pool.
\item For current-query reranking, score eligible resident and completed local blocks with the destination layer's indexer query and select up to 512. For early selection, reuse the nominations for the historical active blocks.
\item Compute attention with the destination attention query, active KV, and eligible local positions.
\end{enumerate}
\end{methodsteps}

\paragraph{Memory scaling.}
With the packed eight-bit KV format (Q8), each token uses 1,088 bytes of KV per sparse layer and four tokens share a 132-byte reference-index entry per layer. Across twelve sparse layers, a million tokens therefore require 13.45\,GB for full KV plus index. With 16,384 or 65,536 KV positions resident per layer, the packed payload for resident KV and the full reference index is about 0.61\,GB or 1.25\,GB, respectively. Appendix~\ref{app:memory} derives these payloads and gives a device example. The full KV backing grows on storage, the index grows in RAM, and the runtime also uses memory for weights, activations, and staging.

\subsection{Layer-major prefill}
\label{sec:layer-major-prefill}
During long prefill, the iPhone can use both native expert routing and native sparse attention by processing a window of prompt tokens through one layer at a time. It loads the current layer's expert weights once per window, and each token's router selects ten experts from all 512. At a sparse attention layer, it loads that layer's historical keys and values, lets each query select blocks across its available causal history, and writes the new cache entries to storage before releasing the layer's KV pages. In both cases, many prompt tokens share the cost of the layer load. During autoregressive decode, each token passes through every layer before the next token is available.

\subsection{Quantization and log-parameterized scale training}
\label{sec:quant}
Q3 and ternary name the quantization formats of the routed expert weights, which account for about 68\% of the model's parameters. Q3 uses a three-bit code per expert weight and ternary uses codes for three values. The complete checkpoint uses 2.54 effective bits per weight for Q3 and 1.66 for ternary. The $n$-gram embedding tables account for another 29\%. Their codes are binary in Q3 and ternary in the ternary model. Most other matrices use four-bit codes. Table~\ref{tab:quant} reports effective bit rates including scale overhead. The complete-model rows include file metadata.

\begin{table}[tbp]
\centering
\caption{Effective bit rates and file sizes for the two Millie 2 checkpoints. Bit rates include scale overhead. Complete checkpoint rates also include file metadata. The format names refer to the routed expert weights.}
\label{tab:quant}
\begin{tabular}{lrr}
\toprule
Component & Q3 & Ternary \\
\midrule
Routed experts (bits/weight) & 3.07 & 1.57 \\
$n$-gram embeddings (bits/weight) & 1.07 & 1.57 \\
Four-bit matrices (bits/weight) & 4.53 & 4.53 \\
Complete model (bits/weight) & 2.54 & 1.66 \\
Model file (GB) & 56.3 & 36.9 \\
\bottomrule
\end{tabular}
\end{table}

\begin{figure}[tbp]
\centering
\includegraphics[width=0.87\linewidth,height=0.39\textheight,keepaspectratio]{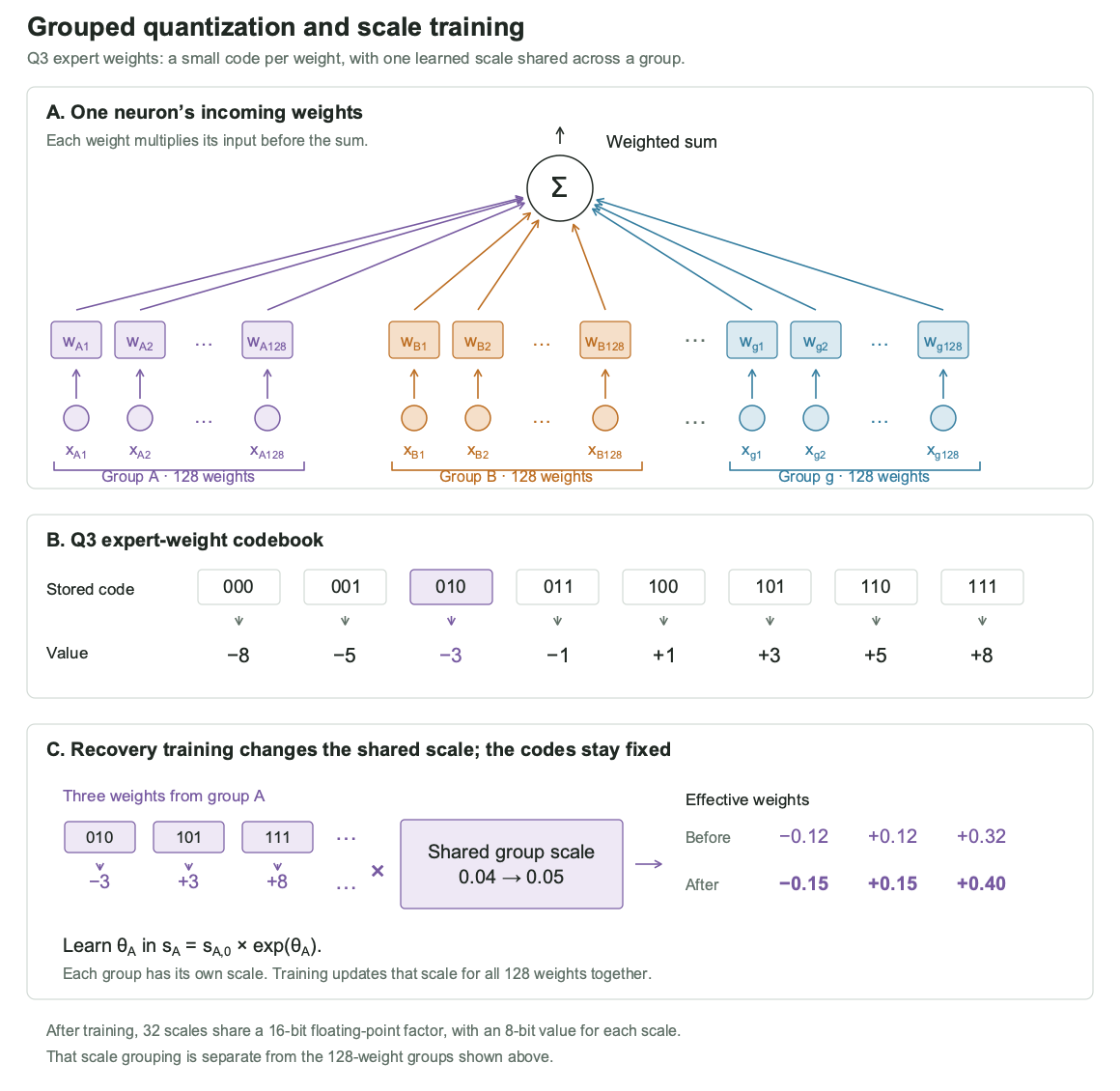}
\caption{Example of grouped quantization for a single expert neuron. The code determines a relative value and the scale determines the magnitude shared across the group. Recovery updates the scale while the codes stay fixed.}
\label{fig:quant}
\end{figure}

\paragraph{Grouped weight representation.}
For an example expert group $g$, write $\widehat w_{g,i}=s_g c_{g,i}$. Q3 expert codes select from $\{-8,-5,-3,-1,1,3,5,8\}$ and ternary codes select from $\{-1,0,1\}$. In ternary experts, half the weights are zero. A nonzero mask and signs for the remaining half use 1.5 bits per expert weight before scale overhead. Routed experts and $n$-gram embeddings use groups of 128 weights. The four-bit matrices use groups of 16. Figure~\ref{fig:quant} illustrates grouping for an expert neuron.

We initialize Q3 expert codes and scales from the parent weights, alternating between assigning each weight to the nearest scaled code and refitting the group scale to minimize squared reconstruction error. For ternary experts, we retain the signs of the 64 largest-magnitude weights in each group, initialize the scale to their mean magnitude, and set the other codes to zero.

\paragraph{Recovery distillation.}
For each quantization group $g$, we keep its integer codes $c_{g,i}$ fixed and train a log multiplier $\theta_g$ for its initial scale $s_{g,0}$. The effective weight is
\begin{equation}
    \widehat{w}_{g,i}(\theta_g)=s_{g,0}\exp(\theta_g)c_{g,i}.
\end{equation}
Specialized kernels read the packed codes during the forward pass and when computing gradients. Initializing $\theta_g=0$ preserves the initial group scale. An update of $\Delta\theta_g$ multiplies every nonzero weight in the group by $\exp(\Delta\theta_g)$, preserving signs and ratios within the group. For routed experts, one trainable scalar thus controls 128 weights. The kernels can reconstruct the required weight values without maintaining a full-precision copy of the model.

To make recovery training more efficient, we use the teacher's next-token logits as targets for the compressed student~\cite{bucila,hinton}. In offline distillation, the teacher generates the completion and supplies the logits at each position. In online distillation, the student generates the completion and the teacher supplies logits conditioned on the student's token history~\cite{imitkd,gkd}.

The student sees the routing restrictions during training, so the scales can adapt to both compressed weights and limited expert availability. We store scales in FP16 during training and compress deployed scales in groups of 32, using a shared FP16 factor and an eight-bit coefficient per scale. Scale-recovery distillation used approximately 39 million completion-token positions for Q3 and 100 million for ternary.

\subsection{Embedding tables and vision input}
At decode time, Millie 2 fetches the few bigram and trigram embedding rows indexed by the current token and keeps the tables on storage. For image prompts, we tested an iPhone runtime that loads the vision encoder weights one layer at a time during image encoding and releases them before text generation.

\section{Evaluation}

\paragraph{Configurations.}
We use the original Qwen3.8-Flash-Next model as the parent reference and as the teacher for distillation. We report Kullback--Leibler (KL) divergence between next-token distributions in nats per token. The main Q3 configuration uses 32 resident experts per layer and an expert admission bias of $+1.5$. Ternary low-memory configurations use 48 residents and the same bias. We run most quality evaluations on GPUs with the corresponding residency and routing restrictions. For the two phone agent evaluations, the model runs on the phone inside the agent loop while a separate machine runs the task environment. Speed and memory measurements use the deployed implementations.

We evaluate two Q3 attention-streaming configurations. The more aggressive setting keeps 16,384 KV positions per sparse layer and gives resident blocks an admission bonus of one standard deviation of the relevance scores. The more conservative setting keeps 65,536 positions and halves the bonus. It retains more of the history in RAM and makes it easier for a highly relevant nonresident block to enter the pool. Appendix~\ref{app:transfers} compares cold loads at each pool size with and without the residency bonus.

\subsection{Reasoning, programming, and instruction following}
Appendix~\ref{app:benchmarks} gives the scoring and run protocols.
The two checkpoint versions are used to answer different questions. For Q3 (Table~\ref{tab:q3quality}), we hold expert streaming fixed and compare attention streaming with native attention. For ternary (Table~\ref{tab:ternquality}), we hold the native attention selection policy fixed and compare the low-memory deployment, which streams experts and uses a Q8 KV cache, with native expert routing and a BF16 KV cache. The ternary comparison measures the combined effect of the expert and cache changes. Both tables also report the original Qwen3.8-Flash-Next model as a reference.

The Q3 streamed runs in this suite use the 16K setting and a 131,072-token total context limit. Prefill uses the streaming policy, and decode applies it once the history exceeds the resident pool. Native attention uses the parent's sparse selection with the full KV cache available. Both HumanEval rows score the same generated functions with different test suites. Both IFBench rows grade the same responses.

\begin{table}[tbp]
\centering
\caption{Effect of Residency-Biased Attention (RBA) on Q3 task scores (\%). GPU evaluation applies the phone's selection policy with simulated KV residency. Both Q3 configurations use the same checkpoint and expert streaming settings. Qwen3.8-Flash-Next is the parent-model reference.}
\label{tab:q3quality}
\begin{tabular}{lrrrr}
\toprule
Benchmark & Tasks & \shortstack{Q3\\RBA} & \shortstack{Q3\\Native attention} & Parent \\
\midrule
GPQA Diamond \cite{gpqa} & 198 & 91.9 & 90.4 & 90.4 \\
HumanEval+ \cite{evalplus} & 164 & 92.1 & 95.1 & 95.7 \\
HumanEval base & 164 & 94.5 & 98.8 & 98.8 \\
IFBench strict \cite{ifbench} & 300 & 81.0 & 83.0 & 82.0 \\
IFBench loose & 300 & 84.0 & 85.3 & 84.0 \\
MuSR \cite{musr} & 756 & 76.6 & 75.9 & 78.8 \\
LiveCodeBench v6 \cite{lcb} & 1,055 & 90.5 & 89.7 & 91.2 \\
\bottomrule
\end{tabular}
\end{table}

The two Q3 configurations score close to the parent on several tasks. Streaming changes the results in both directions across this set. HumanEval exhibits the largest drop with attention streaming among these task scores. The LiveCodeBench comparison uses 955 and 946 successful tasks out of the same 1,055 for streamed and native attention, respectively.

\begin{table}[tbp]
\centering
\caption{Ternary task scores (\%) under a low-memory deployment configuration and an unrestricted baseline. Both use the same ternary checkpoint and native attention selection. Expert routing and KV precision change together. Qwen3.8-Flash-Next is the parent-model reference.}
\label{tab:ternquality}
\begin{tabular}{lrrrr}
\toprule
Benchmark & Tasks & \shortstack{Streamed experts\\Q8 KV} & \shortstack{Native experts\\BF16 KV} & Parent \\
\midrule
GPQA Diamond & 198 & 84.3 & 84.3 & 90.4 \\
HumanEval+ & 164 & 90.2 & 92.7 & 95.7 \\
IFBench strict & 300 & 73.0 & 72.0 & 82.0 \\
IFBench loose & 300 & 76.3 & 75.3 & 84.0 \\
MuSR & 756 & 73.8 & 75.8 & 78.8 \\
LiveCodeBench v6 & 1,055 & 83.3 & 84.3 & 91.2 \\
\bottomrule
\end{tabular}
\end{table}

Across the five suites, giving each suite equal weight and using strict IFBench, the low-memory ternary configuration scores about 0.9 percentage points below the unrestricted ternary configuration. Both ternary configurations score farther from the parent.

\subsection{Agentic evaluations}
Table~\ref{tab:agentserver} reports agent task completion on servers with native attention. Table~\ref{tab:agentphone} reports task completion when Millie 2 runs on the phones with both streaming methods enabled.

\begin{table}[tbp]
\centering
\caption{Server agent task completion (\%). All configurations use native attention. Q3 streams experts, and the two ternary columns report streamed and native expert routing.}
\label{tab:agentserver}
\setlength{\tabcolsep}{4pt}
\begin{tabular}{lrrrrr}
\toprule
Benchmark & Tasks & \shortstack{Q3\\Streamed experts} & \shortstack{Ternary\\Streamed experts} & \shortstack{Ternary\\Native experts} & Parent \\
\midrule
SWE-bench Verified \cite{sweverified} & 500 & 78.4 & 75.2 & 75.2 & 80.6 \\
Terminal-Bench 2.1 \cite{tbench} & 89 & 68.5 & 53.9 & 58.4 & 73.0 \\
\bottomrule
\end{tabular}
\end{table}

SWE-bench Verified tasks run with external network access disabled, and automated tests grade the submitted fixes. A separate run with both expert and attention streaming resolves 387 of 500 tasks (77.4\%), compared with 392 of 500 (78.4\%) with expert streaming and native attention in Table~\ref{tab:agentserver}. Both runs use the release Q3 checkpoint and a 262,144-token context budget. For Terminal-Bench 2.1, we count ungraded first attempts as failures and audit passing trajectories for use of known solutions. In the streamed-expert configurations, Q3 completes 61 of 89 tasks and ternary completes 48. The parent completes 65. The Terminal-Bench runs use an 8,192-token thinking budget, a 262,144-token context limit, and four multi-token prediction (MTP) drafts.

\begin{table}[tbp]
\centering
\caption{Agent task pass rates (\%) with Millie 2 inference on phones and Qwen3.8-Flash-Next inference on a server. Millie 2 uses expert and attention streaming. The phone runs the model that chooses actions, and the separate task harness executes actions and grades the final state.}
\label{tab:agentphone}
\begin{tabular}{lrllrr}
\toprule
Benchmark & Tasks & Checkpoint & Phone & Millie 2 & Parent \\
\midrule
WebArena Lite \cite{webarena} & 70 & Q3 & iPhone 18 Pro & 64.3 & 61.4 \\
AndroidWorld \cite{androidworld} & 116 & Ternary & Samsung Galaxy S26 Ultra & 69.0 & 75.9 \\
\bottomrule
\end{tabular}
\end{table}

On the same 70 WebArena Lite tasks, Q3 running on the iPhone completes 45 and the parent model completes 43. Both complete 39 of the same tasks. The agent keeps its reasoning history and may take up to 30 actions. For AndroidWorld, the model runs on a Samsung Galaxy S26 Ultra and controls apps in a remote emulator. The Android attention policy uses three-layer lookahead and early-query selection.

\subsection{Effect of attention streaming on prediction}
We compare native and streamed attention on an earlier Q3 checkpoint using native prefill in both conditions. We hold the weights, expert configuration, KV formats, and continuation tokens fixed. The streamed condition uses a 16K-token resident pool and a one-standard-deviation residency bias with the earlier lookahead selection policy. We average $D_{\mathrm{KL}}(p_{\mathrm{teacher}}\,\|\,p_{\mathrm{student}})$ across positions within each sequence, then weight sequences equally. For the fixed 16K baseline, we run Q3 with native attention on the most recent input tokens that fit after reserving space for the same fixed continuation in a 16,384-token context. This baseline uses the same checkpoint, expert configuration, KV formats, and KL reference. Across these four groups there are 9,839 continuation positions.

\begin{table}[tbp]
\centering
\caption{Mean teacher-to-Q3 KL divergence in nats/token on fixed continuations. An earlier Q3 checkpoint uses native prefill, then decodes with native attention on the full input, Residency-Biased Attention (RBA) with a simulated 16K-token resident pool on the full input, or native attention on a recent suffix within a 16K-token context budget. Change is the RBA KL minus the native full-input KL.}
\label{tab:attentionkl}
\begin{tabular}{lrrrrr}
\toprule
Context group & Sequences & Native & RBA & Fixed 16K & Change \\
\midrule
CLI, 34K--50K & 4 & 0.00713 & 0.00735 & 0.00888 & $+0.00021$ \\
CLI, 78K--106K & 4 & 0.04940 & 0.05268 & 0.06875 & $+0.00328$ \\
Terminal, 132K--176K & 8 & 0.21209 & 0.21099 & 0.39954 & $-0.00110$ \\
BABILong, about 990K \cite{babilong} & 8 & 0.06487 & 0.07393 & 0.59281 & $+0.00906$ \\
\bottomrule
\end{tabular}
\end{table}

Table~\ref{tab:attentionkl} reports the grouped results. At about 990K, the per-sequence change ranges from about $-0.002$ to $+0.039$ nats/token. All three KL columns measure divergence from the teacher, which includes differences already present under native attention. A negative change means the streamed prediction is closer to the teacher on that sequence. The size of the change varies with the history. On the BABILong continuations, fixed 16K context gives substantially higher divergence than either full-context setting.

\subsection{Question answering at 1M tokens}

We built a shared-context adaptation of BABILong~\cite{babilong} by interleaving renamed bAbI test stories with unrelated book text. Appendix~\ref{app:babilong} describes the construction and evaluation subset. On GPUs, we evaluate 64 questions across four input contexts of approximately 990K tokens each, with supporting facts at least 500K tokens before the questions. Each question is an independent continuation of its context. All configurations use the same Q3 checkpoint with unrestricted expert routing and native sparse attention during prefill. The iPhone can execute this prefill policy layer by layer (Section~\ref{sec:layer-major-prefill}). We vary attention during decode and keep expert streaming enabled. The two streamed settings use one-layer lookahead with current-query reranking. Both rank the reference index for the full history.

\begin{table}[tbp]
\centering
\caption{Question answering on 64 matched questions across four approximately 990K-token BABILong-derived contexts. The same Q3 checkpoint uses native attention and unrestricted expert selection during prefill, and expert streaming during decode in all rows. GPU evaluation simulates KV residency for Residency-Biased Attention (RBA). Resident counts are token positions per sparse-attention layer. Output-limit hits count as failures.}
\label{tab:babilong-qa}
\begin{tabular}{lrr}
\toprule
Decode attention & Correct & Output-limit hits \\
\midrule
Native & 51/64 & 4 \\
RBA, 64K resident, $+0.5\sigma$ & 54/64 & 5 \\
RBA, 16K resident, $+1\sigma$ & 47/64 & 6 \\
\bottomrule
\end{tabular}
\end{table}

In Table~\ref{tab:babilong-qa}, the 64K setting performs on par with native attention on this set (54/64 versus 51/64). The more aggressive 16K setting answers 47 of the same 64 questions correctly. On most benchmarks in Table~\ref{tab:q3quality}, the 16K setting already scores close to native attention, leaving little gap to recover. The 1M-token question-answering result suggests that the more conservative setting may also help on shorter tasks where that gap persists.

\subsection{GPU-to-phone agreement}
To check how well the GPU simulation reproduces phone inference, we run both implementations from a common numerical state and teacher-force the same continuation. Each cell in Table~\ref{tab:parity} uses two trajectories and 4,096 additional positions per trajectory, for 32,768 positions in total. Native-attention prefixes contain 240 or 257 tokens, and streamed-attention prefixes contain 32,767 tokens. With both streaming methods enabled, mean GPU-to-iPhone KL is 0.00581 nats/token for Q3 and 0.00405 for ternary.

\begin{table}[tbp]
\centering
\caption{Mean KL from the GPU distribution to the iPhone distribution in nats/token for fixed continuations from a common numerical state. Expert streaming is enabled throughout. Each cell includes two trajectories and 8,192 continuation positions in total.}
\label{tab:parity}
\begin{tabular}{lrr}
\toprule
Model & Native attention & Streamed attention \\
\midrule
Q3 & 0.00653 & 0.00581 \\
Ternary & 0.00484 & 0.00405 \\
\bottomrule
\end{tabular}
\end{table}

\subsection{Phone throughput, capacity, and power}
For throughput, we use sampled generation with temperature 1, top-$p$ 0.95, and top-$k$ 20. The iPhone 18 Pro uses serial autoregressive decoding. Its Q3 long-document runs use one-layer lookahead with current-query reranking. The REDMAGIC runs use three-layer lookahead and reuse the early selection. The REDMAGIC 12 Pro uses its neural processing unit and up to three MTP drafts \cite{spec}. The reported speeds describe output generation and exclude restoring the prefix and processing the question. A new document first requires prefill to create the saved state. The 1M tests restore the corresponding checkpoint's saved 989,693-token FDA document and process the question before timing generation. The short-context iPhone Q3 $+1.5$ result is the initial 128-token window of a longer response. We use fan cooling for the iPhone $+3$ and FDA runs and for sampled REDMAGIC MTP runs. Table~\ref{tab:phone-speed} reports the fastest measured run for each listed configuration.

\begin{table}[tbp]
\centering
\caption{Sampled answer-generation throughput on phones. Long-context runs restore a saved document prefix of about 990K tokens. Timing starts after the input is processed. Each row gives the fastest measured run for the listed configuration and the number of output tokens measured.}
\label{tab:phone-speed}
\begin{tabular}{llllrr}
\toprule
Model & Phone & Context & Expert bias & Output tokens & Tokens/s \\
\midrule
Q3 & iPhone 18 Pro & Short & $+1.5$ & 128 & 14.1 \\
Q3 & iPhone 18 Pro & Short & $+3$ & 256 & 15.5 \\
Q3 & iPhone 18 Pro & 1M & $+1.5$ & 274 & 9.9 \\
Q3 & iPhone 18 Pro & 1M & $+3$ & 304 & 10.8 \\
Q3 & REDMAGIC 12 Pro & Short & $+1.5$ & 512 & 9.6 \\
Q3 & REDMAGIC 12 Pro & 1M & $+1.5$ & 128 & 7.2 \\
Ternary & iPhone 18 Pro & Short & $+1.5$ & 87 & 16.9 \\
Ternary & iPhone 18 Pro & 1M & $+1.5$ & 52 & 11.7 \\
Ternary & REDMAGIC 12 Pro & Short & $+1.5$ & 512 & 14.3 \\
Ternary & REDMAGIC 12 Pro & 1M & $+1.5$ & 247 & 8.5 \\
\bottomrule
\end{tabular}
\end{table}

REDMAGIC Q3 reaches 6.9 tokens/s over a longer 256-token response to the same FDA question. In the REDMAGIC comparison, both models use the Android attention policy and queued reads, and ternary scores the reference keys on the CPU. On iPhone, Q3 at $+1.5$ averages 8.84 tokens/s across 8,192 generated tokens, with 13.73 tokens/s over the first 512 tokens. The sustained REDMAGIC Q3 run averages 8.68 tokens/s across 2,048 tokens. An iPhone Q3 prefill measurement after the saved 989,693-token prefix processes a further 16,384-token window at 60.6 tokens/s.

\begin{table}[tbp]
\centering
\caption{Q3 context capacity and decode speed on phones. Both phones reach the model's supported length of 1M tokens using Residency-Biased Attention (RBA). To measure decode speed beyond that length, we repeated the saved document cache to construct a longer history. All repeated-prefix measurements use the Q3 release model with 65,536 resident KV positions per attention layer and a $+0.5$ attention residency bias.}
\label{tab:capacity}
\begin{tabular}{@{}lrrrrr@{}}
\toprule
 & & & & \multicolumn{2}{c}{Repeated prefix (tok/s)} \\
\cmidrule(l){5-6}
Phone & RAM & \shortstack{Full KV in RAM\\Tested length} & \shortstack{Streamed\\Model maximum} & 2.5M & 4M \\
\midrule
iPhone 18 Pro & 12\,GB & 147,500 & 1M & 7.6 & --- \\
REDMAGIC 12 Pro & 16\,GB & 229,420 & 1M & 3.5 & 1.5 \\
\bottomrule
\end{tabular}
\end{table}

For the full-KV capacity comparison (Table~\ref{tab:capacity}), both phones restore real segments of the FDA cache and generate 64 sampled tokens. The probes retain the recurrent state from the full checkpoint, so they measure decode capacity from restored state. Checks during decode find all K/V pages resident at the tabulated lengths. The iPhone process exits while restoring the next segment tested, at 155,648 prefix tokens. The REDMAGIC capacity probe uses a 32-slot physical expert buffer.

In separate ternary runs using the full FDA document, both phones cited the correct section, 21~CFR 211.170(b). Both answered that the reserve sample must contain at least twice the quantity needed for the required tests, excluding sterility and pyrogen tests. Appendix~\ref{app:fda} reproduces the question and the two answers.

In a one-hour unplugged iPhone ternary run without external cooling, the device generates 29,001 tokens at 8.06 tokens/s including wall time. Battery charge falls from 75\% to 40\%, as reported in five-point steps by the phone. Screen brightness is about 13\% and low-power mode is off.

\subsection{Memory, speed, and quality across configurations}\label{sec:expert-footprint}
We measure 48 combinations of resident expert count and admission bias on a 24\,GB M5 MacBook Air. Each configuration generates up to 512 sampled tokens from each of two prompts of 87 and 1,797 tokens using the release sampling policy. Decode throughput is the total number of generated tokens divided by the combined decode-loop time. Where we repeated a prompt and configuration, we use the fastest complete trial and its memory measurements. RAM denotes the peak charged process footprint during decode. Quality is mean teacher-to-Q3 KL over 1,024 fixed continuation positions, using the full vocabulary. The sweep uses native attention.

\begin{figure}[tbp]
\centering
\includegraphics[width=0.93\linewidth,height=0.46\textheight,keepaspectratio]{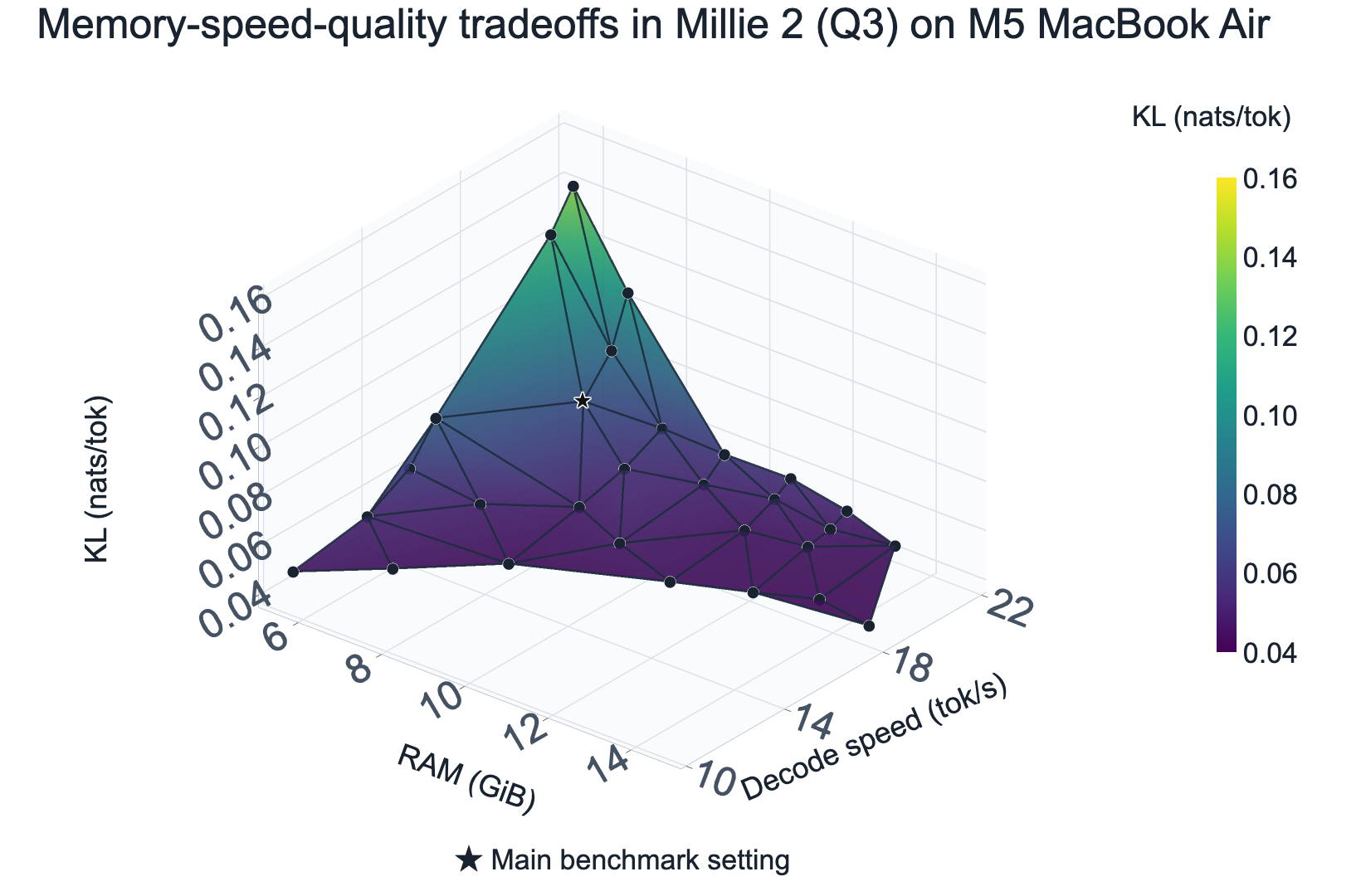}
\caption{Expert-streaming tradeoffs for Q3 on a 24\,GB M5 MacBook Air with short prompts and native attention. RAM is peak decode process footprint. The points are measured settings retained after filtering, and the faces linearly interpolate KL between them. The star marks the 32-resident, $+1.5$-bias benchmark setting.}
\label{fig:frontier}
\end{figure}

Figure~\ref{fig:frontier} shows 29 measured configurations from the sweep. For display we leave out the 16-resident configurations at bias $+2$ and $+3.5$ and filter the remaining points in increasing KL order. A point is omitted if a retained configuration has lower KL, uses at most 0.1\,GiB more RAM, and is at most 0.25 tokens/s slower. We triangulate the retained RAM--speed coordinates and interpolate KL linearly across the faces, omitting triangles that span gaps greater than 3.1\,GiB or 5.1 tokens/s.

The main configuration reaches 20.0 tokens/s at 6.8\,GiB peak decode footprint and 0.071 nats/token. Another measured setting, at 10.9\,GiB, reaches 19.7 tokens/s and 0.047 nats/token. The setting with the lowest plotted KL, 0.045 nats/token, reaches 17.9 tokens/s at 14.9\,GiB. These operating points show that a smaller footprint can preserve most of the measured speed and distributional quality. Residency and bias influence the result differently, and the best setting depends on the available RAM and acceptable change in predictions. A cold expert read delays decoding when it is still pending at the end of the lookahead window. Larger resident pools and stronger bias can reduce how often the reads for a token exceed that window. At 32 resident experts per layer, applying the release bias reduced the median from five cold expert loads to one per token per layer in our GPU evaluation (Appendix~\ref{app:transfers}). On a phone or laptop, a single expert read can often finish while the preceding layer is still computing.

\section{Discussion}\label{sec:discussion}
At 177B parameters, Millie 2 is about five times as large as models previously demonstrated on phones~\cite{millie11}. Even after quantization, the main checkpoint occupies 56.3\,GB on storage. A phone with 12\,GB of RAM runs it at 14 tokens/s at short context and about ten tokens/s when answering a question about a million-token document. The model retains much of its parent's performance across reasoning, coding, instruction-following, and agentic tasks.

This increase in model size changes which models a developer can consider for a phone. A sparse model can have a much larger collection of experts than any one token uses. Its embedding tables can likewise be large while a token needs only the rows it indexes. The practical obstacle is getting those weights into memory by the time the model needs them. Millie 2 predicts which experts to fetch, reuses weights across tokens, and trains under the routing constraints this introduces. A developer can therefore start with a large pretrained model and choose how many of its experts to keep resident on a particular device. As the expert sweep shows, that choice can be made at deployment time with the same checkpoint.

Model size and context length matter together for local applications. An assistant asked why a project decision changed may need to connect an early proposal, a later conversation, and the document that records the decision. RBA lets it search that history while keeping a fraction of the attention data in RAM. In the long-context question-answering experiment, the 64K pool gave scores comparable to native attention after the same native prefill, with about one-eleventh of the RAM for keys, values, and the index (Table~\ref{tab:babilong-qa}). The hardware also has room for models trained for longer histories: in the repeated-prefix test, the iPhone generated at 7.6 tokens/s at 2.5M positions (Table~\ref{tab:capacity}).

The prospect this opens is that personal devices could become a major platform for running large language models. A developer can distribute a model with an application and let the hardware its users already own perform inference. The phone agent evaluations show what this means beyond answering a single prompt: the model can observe the outcome of an action and decide what to do next. When the records and tools it needs are local, that sequence can run privately and continue offline. At the scale of a service, where those decisions run determines what infrastructure the provider needs. Expert streaming could also bring larger models within reach of smaller servers handling a few requests at a time, and RBA could reduce accelerator memory per user on long-context servers. But for an application working with the user's own information, the phone itself is now a place a developer can run a model of this size.

\section{Related work}
\paragraph{Weight offloading and routing.}
LLM in a Flash reduces storage traffic and coalesces reads on memory-limited devices \cite{flash}. Eliseev and Mazur use expert caching and speculative loading, and MoE-Infinity uses activation traces to prefetch and cache experts \cite{eliseev,moeinfinity}. FlashMoE studies SSD-aware expert replacement on edge devices \cite{flashmoe}. Skliar et al. use residency-biased routing for mobile inference \cite{cacheprior}. Our September 2026 Millie 1.1 release describes a 35B phone model with per-layer expert residency, an additive bias for resident experts, and storage reads for selected nonresident experts~\cite{millie11}.

Prefetch-Steered Routing caps predicted admissions to a logical cache, applies its residency bonus at final routing, and issues demand reads for any selected cold experts \cite{psr}. Millie 2 applies the bonus during lookahead admission, gives the strongest cold candidate equal treatment with residents at that stage, and restricts final routing to the resulting resident set. We adapt the compressed model to this restriction through recovery distillation, using the streamed expert policy in the student forward pass.

\paragraph{Sparse attention and KV offloading.}
Native Sparse Attention learns hardware-aligned block sparsity, and DeepSeek-V3.2 uses a lightweight indexer to select tokens for main attention \cite{nsa,dsa}. Quest estimates page relevance from the current query before loading KV data \cite{quest}. HiSparse separates a logical sparse cache from a bounded device-resident cache in a GPU--host memory hierarchy, preserving the indexer's selections \cite{hisparse}. SparDA predicts the next layer's KV blocks with a forecast projection, reuses selected blocks already on the GPU, and overlaps the remaining transfers with computation~\cite{sparda}. In RBA, residency changes the admission scores themselves, so two blocks with equal predicted relevance can receive different priority. We reuse the parent model's block index and put full KV on phone storage. Earlier-layer scores initiate transfers before the destination query is available, and the reranking policy uses that query to choose the blocks consumed by attention.

\paragraph{Low-bit adaptation.}
Quantization-aware training accounts for the deployed weight representation in the forward computation~\cite{jacobqat}. PEQA trains quantization scales with fixed low-bit weight codes. Its main formulation uses per-channel scales, and it also evaluates scales for groups of weights within a channel \cite{peqa}. EfficientQAT ends its training pipeline with end-to-end optimization of quantization parameters \cite{efficientqat}. BitDistiller combines sub-four-bit quantization and distillation \cite{bitdistiller}. For Millie 2, we train log-parameterized group scales on a 177B checkpoint, retain packed frozen weights through forward and backward computation, and distill under the expert-routing restrictions used at deployment.

\section{Conclusion}
Millie 2 runs a 177-billion-parameter language model on phones and generates answers at million-token context. The main checkpoint occupies 56.3\,GB, and even the compressed attention history for a million tokens exceeds the iPhone's 12\,GB of RAM. What determines whether the phone can run the model is how much of each it needs at the same time and whether it can fetch the rest quickly enough. By changing how the model selects experts and retrieves earlier context, we can meet those requirements on hardware people already own. This gives developers the option to bring a capable model to a user's documents, applications, and working history, with the user retaining control over where inference takes place.

\clearpage
\appendix
\section{Benchmark protocols}
\label{app:benchmarks}

\paragraph{Reasoning, coding, and instruction following.}
Tables~\ref{tab:q3quality} and~\ref{tab:ternquality} report the percentage of tasks passed. GPQA Diamond and MuSR are scored by the selected answer. MuSR pools the murder mystery, object placement, and team allocation tasks, giving each example equal weight. HumanEval base and HumanEval+ execute the same generated solution against the original and expanded tests, respectively. LiveCodeBench uses the code generation tasks in release v6 and counts a task as passed when the generated program passes its tests. IFBench reports prompt-level strict and loose scores. In either case all instructions for a prompt must pass the corresponding checks. The loose check also considers benchmark-defined transformations of the response.

For the streamed Q3 suite, we used temperature 1, top-$p$ 0.95, top-$k$ 20, and a 131,072-token total context limit. This includes the LiveCodeBench run. The streamed policy uses native selection through 16,384 tokens. At the boundary it starts with those positions resident, then uses lookahead admission with a one-standard-deviation residency bonus and current-query active selection. The Q3 native and streamed columns use the same recovery checkpoint and expert budget. The ternary comparison changes expert routing and KV precision together, as described in the main text.

\paragraph{Agent tasks.}
Each score is the number of successful tasks divided by the number of tasks in the corresponding table row. SWE-bench Verified uses the benchmark's automated tests to grade submitted repository patches. Terminal-Bench 2.1 uses one first attempt for each of the same 89 tasks. Generation uses temperature 1, top-$p$ 0.95, and top-$k$ 20, with the thinking and context limits given in the main text. The parent, unrestricted ternary, streamed ternary, and Q3 servers support 2, 3, 4, and 4 concurrent requests, respectively. We review passing trajectories for evidence of access to the exact solution and count confirmed cases as failures. Attempts without a grade also count as failures.

In WebArena Lite, phone inference supplies the agent's next action and the separate harness executes the action and grades the task. Both models are evaluated on the same 70-task subset. An episode can end in success, failure, or at its action or time limit. AndroidWorld grades the resulting emulator state. In the Galaxy run, 80 tasks pass, 35 receive a failing grade, and one reaches the runtime limit without a grade, yielding 80 successes over all 116 tasks.

\paragraph{Fixed-continuation comparisons.}
The attention ablation and GPU-to-phone agreement tests score prescribed continuations. A reference and candidate see the same continuation token IDs, so the comparisons measure differences between next-token distributions along the same trajectory. Teacher-to-model KL for the attention ablation is averaged over positions within each history and then over histories. In the fixed 16K condition, space for the entire scored continuation is reserved before selecting the input suffix.

\subsection{Shared-context BABILong adaptation}
\label{app:babilong}
\paragraph{Construction.}
We draw episodes from the English bAbI en-10k test files used by BABILong. The four task families ask for the location of a person using one supporting fact (qa1), the current location of an object using two supporting facts (qa2), the previous location of an object using three supporting facts (qa3), or the giver, recipient, or object in a transfer (qa5). We interleave the story facts, preserving their order within each episode, with unrelated prose from the PG-19 test split. The four contexts use disjoint groups of books, with 42 books in total. Each context contains stories for 64 questions, 16 from each family. We rename people and objects consistently within each episode so that facts from separate stories do not update the same entity. Location names are retained. The shared context contains the story facts. The question is appended separately and gold answers are used only for grading.

A symbolic check verifies the source answers and identifies at least one supporting fact per question whose removal changes or removes the answer. In the reported subset, annotated supporting facts are about 580K to 966K tokens before the question under the model tokenizer. For example, the fact \emph{John\_C01Q01 travelled to the garden} is inserted about 945K tokens before the question \emph{Where is John\_C01Q01?} The gold answer is \emph{garden}.

\paragraph{Evaluation subset and inference.}
We use the first 16 questions in each of the four contexts, for 64 matched questions in total. The subset contains 14 qa1, 16 qa2, 16 qa3, and 18 qa5 questions. Serialized inputs range from 989,994 to 990,031 tokens, including the chat template and question. We prefill with native attention and full expert access and reuse the common prefix computation within each run. Each question starts from the same context prefix and sees no earlier questions or answers. Decoding uses the attention policy in Table~\ref{tab:babilong-qa}, with temperature 1, top-$p$ 0.95, top-$k$ 20, matched per-question seeds, and a common 49,152-token output cap including reasoning. The streamed runs use lookahead admission and current-query active selection. The runtime allows up to 1,048,576 positions for the input and generation.

\paragraph{Scoring.}
A deterministic, question-aware parser grades the explicit final answer. It normalizes superficial formatting and accepts complete answer forms such as \emph{the garden} or \emph{in the garden} for a location. For entity answers it can resolve unambiguous aliases using that question's entity map. A mismatched explicit identifier remains a different entity. The parser compares the canonical answer with the gold label. Generations that reach the output cap or lack a valid final answer count as incorrect.

\clearpage
\section{Million-token FDA document example}
\label{app:fda}

The document prefix contains FDA material and regulations and is 989,693 tokens long. The question and chat framing add 46 tokens in the iPhone demonstration run and 44 tokens in the REDMAGIC run, giving 989,739 and 989,737 input tokens, respectively. The target runtime restores a prefix computed with the corresponding checkpoint and processes the question before starting the decode timer. All target inference runs on the phone. On REDMAGIC the target attends over the full document history through its reference index and resident KV pool, and the draft head starts its cache at the question suffix.

\paragraph{Question (REDMAGIC wording).}
\begin{quote}
Under the FDA rules for reserve samples of finished drug products, how much must a reserve sample contain, and which tests are excluded from that minimum? Give the section number.
\end{quote}

\paragraph{Ternary response on iPhone 18 Pro.}
In the recorded on-device demonstration run, the model generated 52 tokens at 11.65 output tokens/s with reasoning disabled. The final answer was:

\begin{quote}
\textbf{21 CFR 211.170(b)}

A reserve sample for a finished drug product must consist of \textbf{at least twice the quantity necessary to perform all the required tests}, except those for \textbf{sterility and pyrogens}.
\end{quote}

\paragraph{Ternary response on REDMAGIC 12 Pro.}
The run ended after 247 generated tokens at 8.49 tokens/s, including reasoning tokens. The final answer is reproduced below with its original emphasis. The answer identifies the required quantity and both exclusions in the \href{https://www.ecfr.gov/current/title-21/chapter-I/subchapter-C/part-211/subpart-I/section-211.170}{reference provision}.

\begin{quote}
\textbf{21 CFR 211.170(b)}

A reserve sample of each lot or batch of drug product must consist of \textbf{at least twice the quantity necessary to perform all the required tests}, except those for \textbf{sterility} and \textbf{pyrogens}.
\end{quote}

\clearpage

\section{Memory accounting during Q3 decode}
\label{app:memory}
For one sequence at one million positions, a 16-bit reference representation requires approximately 354\,GB for the 177B text weights and another 24.576\,GB for historical keys and values. Table~\ref{tab:weight-payload} follows these data through the Q3 iPhone decode implementation. Each entry is the packed or uncompressed payload required for the indicated data at that step. We use decimal GB and count the reference index separately from the keys and values used by attention.

\begin{center}
\begin{minipage}{\linewidth}
\captionof{table}{From all weights and full attention history to the Q3 decode working set at 1,000,000 positions. The final column uses 32 resident experts and 16,384 resident KV positions per layer.}
\label{tab:weight-payload}
\centering
\begin{tabular}{@{}lrrr@{}}
\toprule
 & \shortstack{All resident\\16-bit (GB)} & \shortstack{All resident\\packed (GB)} & \shortstack{Phone streaming\\packed (GB)} \\
\midrule
Text weights & $\sim$354 & 56.055 & 5.463 \\
Historical keys and values & 24.576 & 13.056 & 0.214 \\
Reference keys & 0.768 & 0.396 & 0.396 \\
\midrule
Weights and attention history & $\sim$379 & 69.507 & 6.073 \\
\bottomrule
\end{tabular}
\end{minipage}
\end{center}

\paragraph{Weights: 56.055 to 5.463 GB.}
The 56.3\,GB Q3 file includes 56.055\,GB of packed text-model tensors, a 0.261\,GB vision encoder and file overhead. A text-only decode uses the following weight data:
\begin{itemize}
\item \emph{Routed experts.} The 512 experts at each of 48 layers account for 46.301\,GB. The iPhone allocates packed weight buffers for 32 experts per layer. Each token selects ten of those 32. Since all experts at a layer have equal size, the buffers hold $46.301\times32/512=2.894$\,GB. The other 43.407\,GB stays on storage, and each new admission reuses a slot.
\item \emph{Input embeddings.} The bigram and trigram tables occupy 6.825\,GB. A token indexes eight rows in each table, each with 160 weights. The ordinary token embedding occupies another 0.360\,GB and supplies one row of 2,560 weights. The phone reads the packed groups covering these rows. Neither table needs a permanent full-table allocation.
\item \emph{Other text weights.} Counting all remaining packed text tensors contributes 2.569\,GB, including the shared experts, attention and recurrent projections, and the 0.360\,GB output projection used to score the vocabulary. Thus the resident experts and remaining text weights together account for $2.894+2.569=5.463$\,GB, apart from transient input rows.
\end{itemize}

\paragraph{Attention: 13.452 to 0.610 GB.}
Each of twelve sparse layers produces two KV heads per token. Each head stores a 256-element key and a 256-element value, giving 1,024 scalars per token per layer. At two bytes per scalar this is 24.576\,GB at one million tokens. Q8 packs 32 scalars into 32 code bytes and a two-byte scale, reducing the row to 1,088 bytes. A separate 128-element reference key summarizes four positions. It occupies 256 bytes at 16 bits or 132 bytes in the packed index. For history length $L$ and $R$ resident positions per layer, the packed payload in RAM is
\[
  12\,R\times1{,}088\quad+\quad12\lfloor L/4\rfloor\times132\quad\text{bytes}.
\]
The first term stops growing when the pool fills. The second continues to cover the full history so the model can nominate a block from anywhere in the input. The full packed KV history remains on storage. Each layer may select up to 512 historical blocks (2,048 tokens) per query from the available data.

\begin{center}
\begin{minipage}{\linewidth}
\captionof{table}{Packed payload in GB at one million positions. Weights are fixed at the 32-resident-expert configuration so the last column isolates the choice of KV pool.}
\label{tab:kv-payload}
\centering
\begin{tabular}{@{}lrrrr@{}}
\toprule
KV positions in RAM per layer & KV & Index & KV + index & With weights \\
\midrule
1,000,000 & 13.056 & 0.396 & 13.452 & 18.915 \\
65,536 & 0.856 & 0.396 & 1.252 & 6.715 \\
16,384 & 0.214 & 0.396 & 0.610 & 6.073 \\
\bottomrule
\end{tabular}
\end{minipage}
\end{center}

\paragraph{From payload to an iPhone decode.}
The 36 recurrent layers also retain a $48\times128\times128$ FP32 state and a $3\times10{,}240$ BF16 convolution history per layer. Including the embedding module's $9\times10{,}240$ BF16 history adds 0.116\,GB in total, bringing the 16K and 64K accounted payloads to 6.189 and 6.830\,GB. The phone also needs working space for activations, selection scores, temporary reads, allocator and GPU buffers, and the application and OS. File-backed pages and device allocations have different physical-memory accounting from sums of tensor sizes. The direct check is generation on the 12\,GB phone: at 2.5M repeated-prefix positions the runtime populated all 65,536 KV slots per layer (0.856\,GB), indexed the full history (0.990\,GB), and generated 128 tokens at 7.6 tokens/s. The corresponding weight, KV, index and recurrent-state payload is 7.424\,GB.

\paragraph{Deployed buffers on REDMAGIC.}
In the repeated-prefix test at 2.5M positions, the REDMAGIC stored resident KV rows in decoded FP16 form, at 2,048 bytes per position per layer. The populated 65,536-position pools therefore contained $12\times65{,}536\times2{,}048=1{,}610{,}612{,}736$ bytes. The 625,000 reference keys per layer added 990,000,000 bytes of index payload, for 2.601\,GB across these two components. At 4M positions, the same pools and a larger index accounted for 3.195\,GB. The runtime also allocates expert buffers (48 physical device slots for the 32-resident routing configuration in these runs), recurrent state, activations, selector and transfer buffers, and the speculative draft head. The payload sums account only for the named components. Section~\ref{sec:expert-footprint} reports whole-process decode footprint for the Mac sweep.

\clearpage
\section{Lookahead from the four-branch residual stream}
\label{app:lookahead}
Both predictors read the residual stream with parameters from the layer that will use the data. The expert predictor reads after sequence mixing in the preceding layer. The attention predictor reads at the entry to an earlier layer. We give the operations for one token at position $t$, omitting rounding and packing.

\paragraph{How a destination module reads the residual.}
Write $H=(h_1,\ldots,h_4)$ for the four residual branches, each of width $p=2560$. Let $m$ identify a particular layer and module (expert or attention). For any vector $v$ of width $n$ and effective learned gain $\gamma$, define
\[
  N_\gamma(v)=\frac{\gamma\odot v}{\sqrt{n^{-1}\sum_{i=1}^n v_i^2+10^{-6}}}.
\]
The module normalizes each branch separately, then uses all four to compute an elementwise gate for each branch. Its read vector $M_m(H)$ is
\begin{align*}
  u_c &= N_{\gamma_{m,c}}(h_c), \qquad u=[u_1;u_2;u_3;u_4],\\
  [g_1;g_2;g_3;g_4] &= \operatorname{sigmoid}\!\left(U_m\,\operatorname{SiLU}(D_m u/4)\right),\\
  M_m(H)&=\tfrac14\sum_{c=1}^4 g_c\odot u_c .
\end{align*}
Here $D_m$ maps the concatenated 10,240 coordinates to 320 and $U_m$ maps them back. Separate parameter sets serve the expert and attention modules of each layer. Applying $M_m$ early is a read: the intervening layers continue updating the residual in their usual order. When the destination module reads again, it uses the same parameters and recomputes the gates from the updated branches.

\paragraph{Expert lookahead.}
Let $H_i^{\rm in}$ denote the residual at entry to layer $i$'s sequence mixer and $H_i^{\rm mid}$ the residual after its output has been incorporated, before the experts run. Write $E_\ell$ for the expert-input mixer of destination layer $\ell$. For layers with a preceding layer, the two inputs and the two router outputs are
\begin{align*}
 x^- &= M_{E_\ell}(H_{\ell-1}^{\rm mid}), & a &= W_\ell x^- + B_\ell A_\ell x^-,\\
 x   &= M_{E_\ell}(H_\ell^{\rm mid}),     & z &= W_\ell x .
\end{align*}
The early read precedes both the experts of layer $\ell-1$ and the sequence mixer of layer $\ell$. The LoRA adapter $B_\ell A_\ell$ supplies a correction only to the early prediction. Let $R$ be the resident experts and $P$ the protected nonresident candidates defined in Section~\ref{sec:expert}. The nominations $N$ and active experts $S$ are
\[
 N=\operatorname{Top}_{10}\!\left((a_e+b\,\mathbf{1}\{e\in R\cup P\})_e\right),
 \qquad
 S=\operatorname{Top}_{10}\!\left((z_e)_{e\in R'}\right),
\]
where $R'$ is the pool after admitting $N$. The current router supplies the mixture weights for $S$. A resident expert can enter $S$ without having been nominated on this token.

\clearpage
\paragraph{Attention lookahead.}
For a sparse destination layer $\ell$, write $T_\ell$ for its attention-input mixer. The early read comes from layer entry, before sequence mixing or expert computation there. With lookahead distance $d$, the early and current inputs are
\[
 x^- = M_{T_\ell}(H_{\ell-d}^{\rm in}),\qquad
 x = M_{T_\ell}(H_\ell^{\rm in}).
\]
For indexer head $a=1,\ldots,4$, let $Q_{\ell,a}$ be its 128 query rows of the destination indexer's query--key projection. Both reads use those same rows. For an input $y$ and a stored reference key $r_{\ell,j}$ of completed four-token block $j$, the query and block score are
\begin{align*}
 q_a(y)&=\operatorname{RoPE}_t\!\left(N_{\gamma_q}(Q_{\ell,a}y)\right),\\
 s_j(y)&=\frac{1}{\sqrt{128}}\sum_{a=1}^4
             \max\!\left(0,\,q_a(y)^\top r_{\ell,j}\right).
\end{align*}
The indexer rotates the first 64 coordinates at the current token position. Its reference key is built once per block: for a block starting at position $u$, the indexer averages the four raw index keys, normalizes the result, and applies the rotary transform at $u$. Each reference key can therefore be scored before the corresponding per-token keys and values are in RAM.

Let $I$ be the blocks completed before token $t$, $R$ the resident block IDs, and $\sigma^-$ the standard deviation of $s_j(x^-)$ over $j\in I$. During streaming, the nominations are
\[
 N=\operatorname{Top}_{512}\!\left(
     (s_j(x^-)+\beta\sigma^-\mathbf{1}\{j\in R\})_{j\in I}\right).
\]
Reads for missing nominees begin while earlier layers compute. Let $C$ contain the eligible completed blocks after admission, including a block just completed locally when applicable. The historical selection for the two decode policies is
\begin{align*}
 S_{\rm iPhone}  &= \operatorname{Top}_{512}\!\left((s_j(x))_{j\in C}\right) && (d=1),\\
 S_{\rm Android} &= N && (d=3,\ \text{early selection}).
\end{align*}
The iPhone recomputes the indexer queries from $x$ and ranks eligible blocks without a residency bonus. The Android early-selection policy reuses its nominations. In both cases, the destination layer computes its ordinary attention query from its current input $x$ and attends to individual token keys and values in the selected blocks and the eligible local tail. The reference-key dot products above select blocks; they are distinct from the per-token attention scores.

\section{Cold loads during decoding}
\label{app:transfers}

We measured how often the release Q3 model needed to fetch expert weights or historical key-value (KV) data during decoding. In each GPU comparison, we kept the resident pool size and lookahead policy fixed and varied the residency bias. Each arm sampled up to 256 output tokens on each of four shared prompts. We count an admission of nonresident data as one cold load. These counts simulate transfer demand from the runtime's selection policy. Flash transactions and wait time also depend on how the runtime batches and schedules those loads.

\begin{figure}[htbp]
\centering
\includegraphics[width=.8\linewidth]{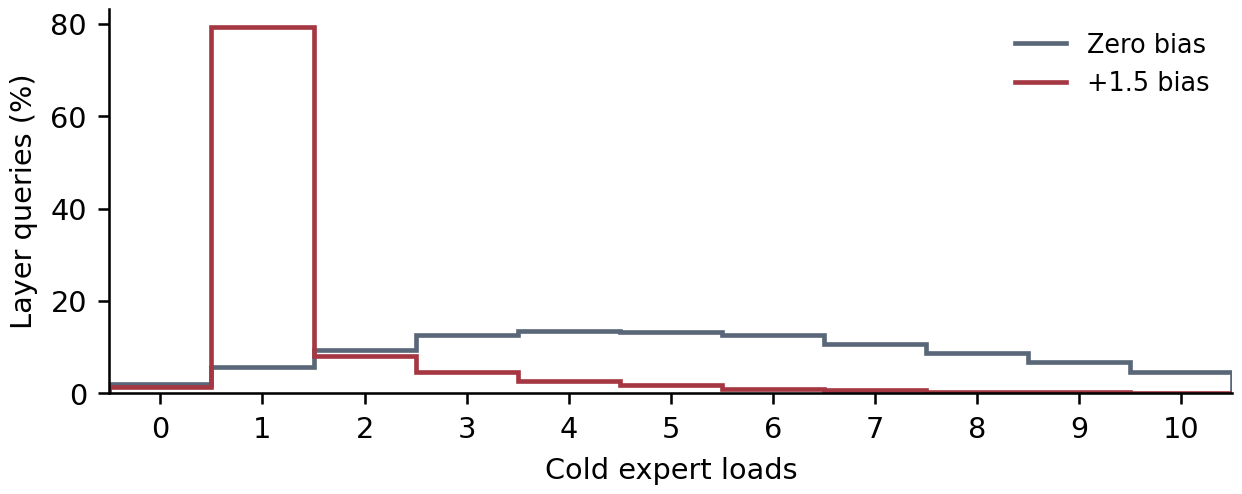}
\caption{Cold expert loads per token per expert layer with native attention. Each arm contributes 48{,}960 layer queries.}
\label{fig:expert-transfers}
\end{figure}

\paragraph{Experts.}
At 102{,}400 input tokens, we used native attention throughout and kept 32 experts resident per layer during decoding. The release bias of $+1.5$ reduced the mean number of cold expert loads per token per layer from 5.10 to 1.45. The fraction of token-layer steps with more than one cold expert load fell from 92.2\% to 19.4\%. Figure~\ref{fig:expert-transfers} shows the full distribution.

\begin{figure}[htbp]
\centering
\includegraphics[width=\linewidth]{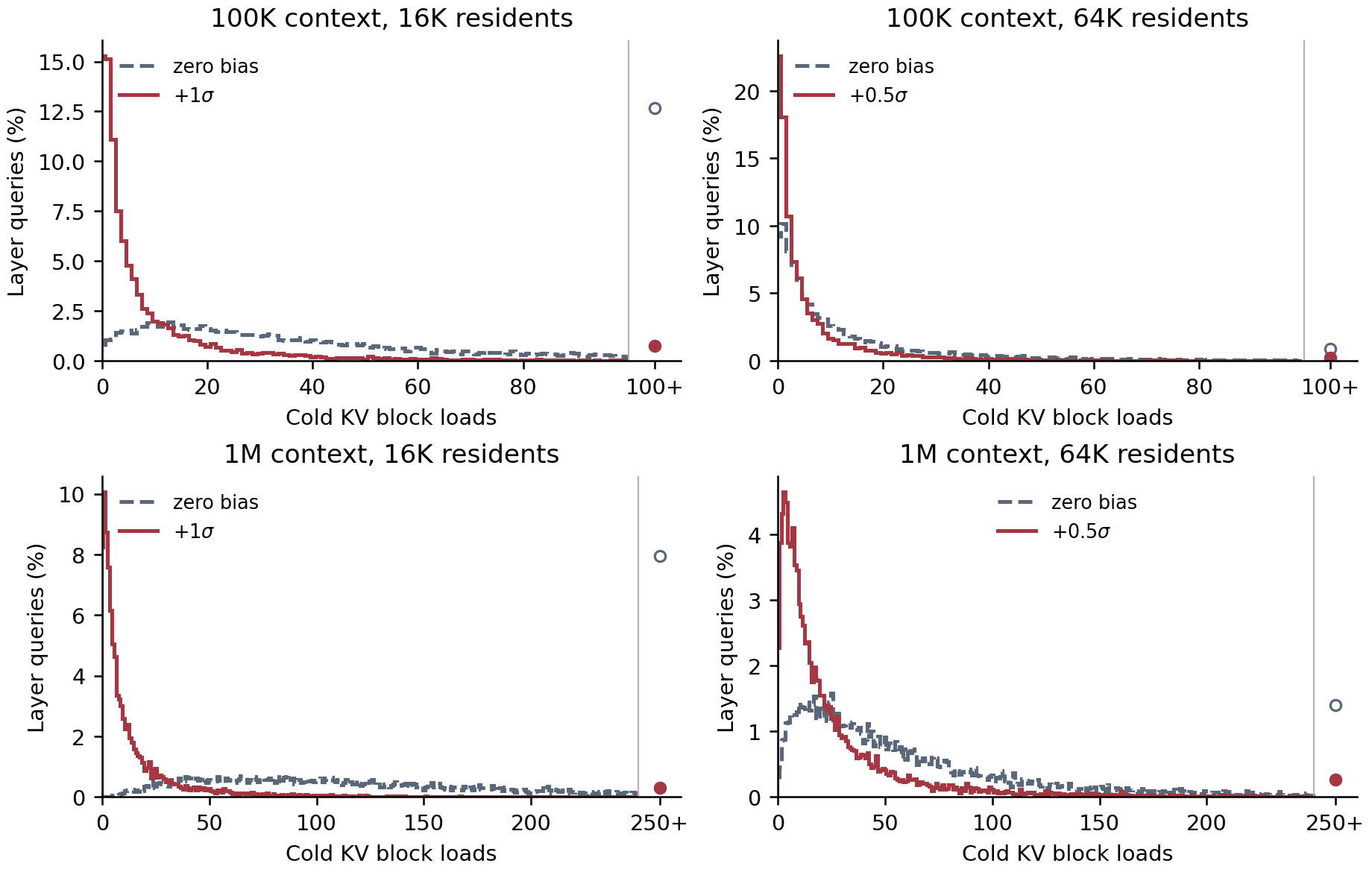}
\caption{Cold KV block loads per token per sparse attention layer. Each panel compares the configured bias with zero bias at the same pool size. Vertical scales vary by panel. The separate markers at the right collect all counts at or above the labeled cutoff.}
\label{fig:attention-transfers}
\end{figure}

\paragraph{Attention.}
We compared the two RBA configurations at 102{,}400 and 989{,}968 prefix tokens. All arms used native attention during prefill and one-layer lookahead with current-query reranking during decode. Each histogram in Figure~\ref{fig:attention-transfers} shows the number of newly admitted four-token KV blocks per token per sparse attention layer. At 100K, the mean fell from 50.62 to 10.27 for the 16K pool and from 14.31 to 6.37 for the 64K pool. At 1M, it fell from 123.78 to 15.93 and from 60.57 to 26.05, respectively. Each arm contributes 12{,}240 layer queries, except the 1M zero-bias 16K arm, which ended earlier and contributed 10{,}656.

\paragraph{When transfers affect latency.}
Lookahead lets the runtime issue reads while earlier layers are still computing. Reads that finish before the destination layer needs the data add little or no delay to decoding. The amount of traffic that fits in this window depends on the device's storage bandwidth, read latency, concurrency, and layer compute time.

\end{document}